%% file: colm2026_conference.tex
\documentclass{article} % For LaTeX2e
\usepackage[preprint]{colm2026_conference}

\usepackage{microtype}
\usepackage{hyperref}
\usepackage{url}
\usepackage{booktabs}
\usepackage{multirow}
\usepackage{graphicx}
\usepackage{subfloat}
\usepackage{booktabs}
\usepackage{subfig}
\usepackage{amsmath}
\usepackage{subcaption}
\usepackage[normalem]{ulem}
\usepackage{algorithm,algpseudocode}
\algrenewcommand\textproc{}
\usepackage{amsmath}
\usepackage{amssymb}
\usepackage{xcolor}
\usepackage{lineno}
\usepackage{color}
\usepackage{colortbl}
\usepackage{amsfonts}
\usepackage{makecell}
\usepackage{paracol}
\usepackage{listings}
\usepackage{wrapfig}
\usepackage{seqsplit}
\usepackage{marvosym} 
\usepackage{textcomp}
\usepackage{enumitem}

\definecolor{myred}{RGB}{220,50,47}    % A slightly muted, sophisticated red
\definecolor{mygreen}{RGB}{133,153,0}  % A subdued, olive-like green
\definecolor{darkblue}{rgb}{0, 0, 0.5}
\hypersetup{colorlinks=true, citecolor=darkblue, linkcolor=darkblue, urlcolor=darkblue}

\title{Counting and Min-Cost Encoding for Tokenization in Large Language Models}

\author{Shuming Shi$^{1}$, Xiang Zhang$^{1}$, Hao Yu$^{2}$, Wenbo Fei$^{2}$, Changjian Wang$^{1}$, \\
\textbf{Zhan Wang}$^{2}$, \textbf{Guoqing Pang}$^{2}$, \textbf{Guangye Yu}$^{1}$, \textbf{Quan Lu}$^{1}$, \textbf{Ning Jiang}$^{1}$\\
$^{1}$Mashang Consumer Finance Co., Ltd., China\quad \\
$^{2}$National-Mathematics Artificial Intelligence Institute in Chongqing (NMAII), China \quad \\
}

\begin{document}

\ifcolmsubmission
\linenumbers
\fi

\maketitle

\begin{abstract}
    \input{sections/0_abstract}

\end{abstract}

%%% PAPER SECTIONS %%%

\input{sections/1_introduction}
\input{sections/2_related_work}
\input{sections/3_method}

\input{sections/4_experiments_n}

\input{sections/5_conclusion}

\bibliography{colm2026_conference}
\bibliographystyle{colm2026_conference}

\appendix
\input{sections/appendix}

\end{document}

%% file: sections/0_abstract.tex
Mainstream large language models (LLMs) rely on a tokenizer to encode text into a token sequence. Different tokenizers may yield token sequences of substantially different lengths for the same text. With a fixed model architecture, shorter token sequences correspond to lower inference time.
We propose a tokenizer training approach named Counting and Filtering (CNF) and a text encoding algorithm called Min-Cost Encoding (MCE).
MCE defines a cost function over a text segment, and determines the best segmentation by globally minimizing the overall segmentation cost.
CNF builds a raw vocabulary by directly counting valid substrings, and then constructs the final vocabulary through a filtering step based on actual token usage when segmenting the training corpus with MCE.
The CNF-MCE conbination offers several advantages over BPE (the most widely used tokenizer algorithm for LLMs so far), including higher token efficiency, greater scalability, and lower dependency.
Across six text categories and two vocabulary-size groups, CNF-MCE consistently achieves better compression than the evaluated BPE tokenizers.
With a 250K vocabulary, CNF-MCE increases compression rate by 26\% and 30\% on English web text over the o200k\_base and qwen250k tokenizers.
Experiments scaling the vocabulary to 1M entries on English web text demonstrate sustained improvements over BPE, with a token efficiency improvement of over 60\% and vocabulary utilization rising from 52.9\% to 96.9\%.
The MCE algorithm does not depend on a merge list (as in BPE) or token probability (as in UnigramLM), making it applicable to a wide range of vocabularies, including those built from BPE, UnigramLM, CNF, and others.
Language models trained from scratch at the 1.8B and 8B scales achieve comparable average performance to models using the BPE tokenizers across 11 benchmarks. These results demonstrate that CNF-MCE can improve token efficiency significantly while maintaining competitive downstream performance.

%% file: sections/1_introduction.tex
\section{Introduction}

Transformer-based large language models (LLMs) cannot process raw text directly. They rely on a tokenizer to convert input text to a sequence of tokens, a process referred to as text encoding, tokenization, or segmentation. Similarly, the output token IDs produced by LLMs are converted back into text through a decoding process. A tokenizer is primarily defined by a vocabulary, an encoding algorithm, and a decoding algorithm.
The choice of tokenizers directly affects both the efficiency and performance of LLMs. For the same text, different tokenizers may produce token sequences of substantially different lengths.
With a fixed model architecture, shorter token sequences correspond to lower inference time.
Token efficiency is commonly quantified by the \emph{compression rate}, expressed as the average amount of raw text (e.g., UTF-8 bytes) encoded per token; a higher rate therefore indicates a shorter token sequence \citep{meister2026tokeval,liu2025superbpe}.
An important objective in tokenizer design is therefore to improve compression rate under a fixed vocabulary budget while preserving or improving downstream model performance.

Widely used tokenizer training methods face structural limitations in optimizing compression under a fixed vocabulary budget. Byte Pair Encoding (BPE) constructs its vocabulary through a sequence of order-dependent merges \citep{sennrich2016neural}. Forming a long token requires introducing intermediate merge products, some of which may remain underutilized while continuing to consume vocabulary capacity \citep{chizhov2024pickybpe}. Hard pretokenization boundaries further prevent tokens from spanning multiple pretokens. Recent methods such as SuperBPE \citep{liu2025superbpe} relax this boundary constraint but retain sequential merge-based construction. UnigramLM avoids merge ancestry, but it requires probability estimation over a segmentation lattice and iterative likelihood-based pruning \citep{kudo2018subword,land2026mingram}. These approaches therefore tie vocabulary construction to either sequential merge operations or iterative  refinement. This raises a natural question:
can we significantly improve token efficiency under a fixed vocabulary budget without relying on merges or repeated vocabulary refinement?

To address this question, we propose a tokenizer training approach named \textbf{Counting and Filtering} (\textbf{CNF}) and a text encoding algorithm called \textbf{Min-Cost Encoding} (\textbf{MCE}).
In MCE, a cost function is defined for a segment in a piece of text, and the best segmentation is determined by globally minimizing the overall segmentation cost.
Our cost function does not rely on merge information (as in BPE) or token probability estimation (as in UnigramLM), making MCE applicable to a wide range of vocabularies.
CNF consists of two phases. In the counting phase, a raw vocabulary is built by calculating the occurrences of valid substring candidates in the training corpus and selecting the $m$ most frequent ones. In the filtering phase, the training corpus is first segmented by MCE based on the raw vocabulary. Then the segmentation frequency of each token is calculated and the final vocabulary is constructed by filtering out from the raw vocabulary the tokens with low segmentation frequencies.
Note that CNF encodes the training corpus only once, which makes the filtering step very efficient.
This non-iterative CNF procedure shifts vocabulary selection from raw occurrence frequency to realized usage under MCE, yielding a vocabulary aligned with actual token usage.

We evaluate our method using intrinsic tokenizer metrics across multiple text domains and vocabulary sizes, as well as downstream task performance.
Our method achieves consistently better compression than the evaluated BPE tokenizers.
Particularly, the MCE encoding algorithm on our mix250k vocabulary achieves 26.7\% and 32.4\% higher token efficiency than the o200k\_base and qwen250k tokenizers respectively.
Experiments scaling the vocabulary to 1M entries on English web text demonstrate sustained improvements over BPE, with a token efficiency improvement of over 60\% and vocabulary utilization rising from 52.9\% to 96.9\%.
Language models trained with our tokenizers at the 1.8B and 8B scales achieve comparable average downstream performance across 11 English benchmarks to models trained with the BPE tokenizers.

Our main contributions are as follows:
\begin{itemize}
    \item We introduce CNF, a merge-free vocabulary construction method that builds a raw vocabulary by counting valid substring occurrences and selects the final vocabulary through a single filtering step based on realized token usage under MCE.
    \item We propose MCE, a segmentation algorithm that globally minimizes an encoding cost that combines token count with penalties for partial-word boundary crossings. MCE can be applied to a wide range of vocabularies, including those built from BPE, UnigramLM, CNF, and other methods.
    \item We demonstrate improved compression and a range of intrinsic tokenizer properties, sustained compression gains with high vocabulary utilization as vocabulary size increases, and comparable average downstream performance to the evaluated baselines.
\end{itemize}

%% file: sections/2_related_work.tex
\section{Related work}
\label{sec:relatedwork}

\paragraph{Subword tokenization.}
BPE \citep{sennrich2016neural} and UnigramLM \citep{kudo2018subword} represent two widely used approaches to subword tokenization. BPE builds a vocabulary by repeatedly merging the most frequent adjacent symbol pair, and encodes new text according to the learned merge order. Its intermediate merge products occupy vocabulary entries, although some are rarely used in final segmentations. PickyBPE \citep{chizhov2024pickybpe} and Scaffold-BPE \citep{lian2025scaffoldbpe} address this issue by removing such tokens based on relative frequency criteria during training. UnigramLM instead starts from an overcomplete candidate vocabulary, estimates token probabilities with EM, and iteratively prunes tokens according to their contribution to corpus likelihood, using Viterbi decoding to select the most probable segmentation. MinGram \citep{land2026mingram} simplifies this procedure through BPE-based initialization, Hard EM on minimum-token paths, and a single final score-based pruning step. Vocabulary construction and segmentation can also be studied separately: \citet{uzan2024greed} compare inference methods across multiple vocabulary construction algorithms, while PathPiece \citep{schmidt2024tokenization} combines minimum-token segmentation with iterative vocabulary pruning and supports multiple initialization strategies. 
Our framework constructs the vocabulary through direct substring counting and a single filtering step based on MCE token usage, without sequential merges or iterative refinement.
Moreover, unlike previous methods that impose hard word-level pretokenization boundaries, our framework supports multiword tokens, allowing for further sequence compression.

\paragraph{Tokenization beyond word boundaries.}
Many subword tokenizers impose hard pretokenization boundaries, typically based on whitespace and punctuation, which prevent tokens from spanning multiple words \citep{liu2025superbpe,schmidt2025boundlessbpe}. Early work explored incorporating statistically selected multiword expressions into tokenization, with evaluations primarily focused on machine translation and text classification \citep{kumar2022bpe,gee2023multiword}. More recently, SuperBPE \citep{liu2025superbpe} introduces a pretokenization curriculum that first learns subwords and then permits merges across whitespace, demonstrating improvements in both compression and downstream performance through controlled language-model pretraining experiments. BoundlessBPE \citep{schmidt2025boundlessbpe} selectively merges adjacent pretokens that are each represented by a single token into superwords, preserving complete pretoken boundaries during these merges. Although these methods support multiword tokens, their vocabulary construction and inference still rely on ordered merge operations. Our approach instead controls partial-word boundary crossings through candidate validity rules during CNF vocabulary construction and through penalties and constraints during MCE segmentation, without relying on ordered merge operations.

%% file: sections/3_method.tex
\section{Methodology}
\label{sec:method}

A tokenizer is defined mainly by three elements: a vocabulary, an encoding algorithm, and a decoding algorithm.
Our approach comprises two algorithms: \textbf{Counting and Filtering} (\textbf{CNF}) for vocabulary construction, and \textbf{Min-Cost Encoding} (\textbf{MCE}) for text tokenization.
Decoding reduces to a simple table lookup in the vocabulary's (ID, token text) dictionary.
CNF first constructs a raw vocabulary through substring counting in the training corpus and then refines it based on token usage under MCE. Given a vocabulary, MCE selects a segmentation by minimizing an encoding cost.

\subsection{Vocabulary Construction: Counting and Filtering (CNF)}
\label{sec:cnf}

CNF consists of two phases. In the counting phase, we construct a raw vocabulary by selecting the most frequent candidate tokens in the corpus. In the filtering phase, we construct the final vocabulary by removing tokens from the raw vocabulary that are infrequently used when MCE tokenizes the corpus. Thus, counting measures candidate occurrences in the original text, whereas filtering measures actual token usage during tokenization.

\subsubsection{Raw Vocabulary Construction by Counting}
\label{sec:counting}

Given a text corpus $C$, we extract valid candidate tokens from each document and aggregate their occurrence counts across the corpus. Then a raw vocabulary is built by selecting the top-$m$ candidates with the highest counts.

A critical step in the counting phase is determining which substrings qualify as valid candidate tokens. BPE vocabulary construction typically begins with pretokenization, which splits text into chunks based on delimiters (e.g. whitespace) and other rules. Merges are restricted to individual chunks, preventing tokens from spanning chunk boundaries. CNF relaxes this boundary restriction to allow words, subwords, and multiword phrases as candidate tokens. This expands the candidate space but may also introduce undesirable cross-word substrings, such as those that begin within a word. For example, consider the following text:

  \begin{center} Example text: "1600 B.C. was a significant year in human history." \end{center}

Under CNF's validity rules, ``6'', ``\textvisiblespace sig'', ``significant'', ``cant'', and ``\textvisiblespace year'' are  valid candidate tokens. Multiword phrases such as ``\textvisiblespace was\textvisiblespace a'' and ``significant\textvisiblespace year'' are also valid. However, ``ficant\textvisiblespace year'' is excluded because  it spans multiple words but begins within a word. Given a text $T$ and a substring $S$, we determine the validity  of $S$ using rules that account for its boundaries and context  within $T$.

Although validity checks eliminate many substrings, a document can still yield a large number of candidate tokens, making exhaustive counting costly for corpora containing millions of documents. We address this challenge with an efficient counting algorithm for raw vocabulary construction, summarized in Algorithm~\ref{alg:efficient-counting}. The algorithm uses two main strategies:

\paragraph{Multi-round counting with early stopping.}
We construct the raw vocabulary in multiple rounds over consecutive, non-overlapping length intervals starting at length one, with shorter candidates processed first. Each interval $r=[r_{\min},r_{\max}]\in R$ has an associated frequency threshold $t_r$. 
At the start of each round, we build a byte trie from all candidates retained in earlier rounds, storing their corpus-wide frequencies for comparison with the current threshold $t_r$ during enumeration. The trie is empty in the first round.

The \textsc{CountCandidates} procedure enumerates substrings by starting position and increasing length, using two early-stopping criteria. First, if a substring is invalid and cannot become valid through further extension, enumeration stops at that starting position (e.g., in ``a significant year in human history'', once the substring ``ficant\textvisiblespace'' is reached, further extension cannot resolve its invalid start within a word). Invalid substrings that may become valid through extension are skipped without stopping enumeration. Second, if a valid substring within a previously processed length interval is absent from the trie or has a stored frequency below the current threshold $t_r$, enumeration stops at that starting position. In either case, the prefix and its extensions cannot meet the current frequency threshold. Valid candidates within the current round's length interval are counted.

These criteria reduce unnecessary substring enumeration and the number of candidates stored in frequency dictionaries, reducing computation time and memory usage.

\paragraph{Parallel counting and aggregation.}
Within each round, workers process disjoint subsets of documents in parallel, each maintaining a local frequency dictionary. We merge these dictionaries by summing the counts of identical candidates across workers, then retain candidates with corpus-wide frequencies of at least the current threshold $t_r$. After all rounds, we select the top-$m$ candidates by frequency from the combined set of retained candidates to form the raw vocabulary.

In our experiments, it takes less than one hour to run the raw vocabulary construction algorithm (using 32 threads) on a corpus of one million English documents.

\begin{algorithm}[t]
\caption{Raw Vocabulary Construction by Efficient Counting}
\label{alg:efficient-counting}
\begin{algorithmic}[1]
\Require Corpus $C$; ordered length intervals $R$;
frequency thresholds $\{t_r\}_{r \in R}$; raw vocabulary size $m$
\Ensure Raw vocabulary $V_0$
\State $F \gets \emptyset$ \Comment{Retained candidates and their counts}
\For{$r \in R$}
    \State $\mathcal{T} \gets \Call{BuildByteTrie}{F}$ \Comment{Store candidate counts}
    \For{each document $d \in C$ \textbf{in parallel}}
        \State $f_d \gets \Call{CountCandidates}{d,r,\mathcal{T},t_r}$
    \EndFor
    \State $f \gets \sum_{d \in C} f_d$ \Comment{Sum counts by candidate}
    \State $F \gets F \cup \{(s,f(s)) : f(s) \ge t_r\}$
\EndFor
\State \Return the top-$m$ candidates in $F$ ranked by frequency
\Statex
\Procedure{CountCandidates}{$d,r,\mathcal{T},t$}
    \State $f \gets$ an empty frequency dictionary with default value $0$
    \For{$i = 0,\ldots,|d|-1$}
        \For{$\ell = 1,\ldots,\min(r_{\max},|d|-i)$}
            \State $s \gets d[i:i+\ell]$
            \If{$s$ is invalid at position $i$ in $d$}
                \If{no extension of $s$ can become valid}
                    \State \textbf{break} \Comment{Stop extending this prefix}
                \EndIf
                \State \textbf{continue} \Comment{Skip counting but allow extension}
            \EndIf
            \If{$\ell \ge r_{\min}$}
                \State $f(s) \gets f(s)+1$
            \ElsIf{$s \notin \mathcal{T}$ \textbf{or} $\mathcal{T}(s) < t$}
                \State \textbf{break} \Comment{Missing or low-frequency prefix}
            \EndIf
        \EndFor
    \EndFor
    \State \Return $f$
\EndProcedure
\end{algorithmic}
\end{algorithm}

\subsubsection{Filtering}
\label{sec:filtering}

The counting phase assigns each candidate token its substring  frequency in the training corpus. However, this frequency does not necessarily reflect how often the token is used in tokenization. For example, ``\textvisiblespace significan'' occurs in both ``\textvisiblespace significant'' and ``\textvisiblespace significance'', and thus has a substring frequency at least as high as that of either word. But, when both complete words are included in the vocabulary, this shared prefix  may be selected infrequently during tokenization.

Such candidates resemble BPE ``junk'' tokens, i.e., intermediate tokens  that participate in vocabulary construction but rarely appear in the resulting tokenized text \citep{bostrom2020suboptimal,chizhov2024pickybpe}. These tokens occupy  vocabulary slots despite their limited use and may receive insufficient training in downstream language models \citep{chizhov2024pickybpe,land2024magikarp}. 
This motivates us to identify and filter out underused candidates by measuring their frequencies in the tokenization results of a sufficiently large text collection.

In the filtering phase, we first apply MCE tokenization (Section~\ref{sec:mce}) to the training corpus using the raw vocabulary obtained in the counting phase. We then compute each token's \emph{segmentation frequency}, defined as its number of occurrences in the resulting token sequences. The final vocabulary consists of the top-$k$ tokens ranked by segmentation frequency.

\subsection{Text Tokenization: Min-Cost Encoding (MCE)}
\label{sec:mce}

Given a fixed vocabulary, MCE selects a segmentation by globally minimizing an encoding cost. This cost combines token count with penalties for partial-word boundary crossings and an optional token preference term. MCE does not depend on how the vocabulary was constructed and can therefore operate on the raw CNF vocabulary, the final CNF vocabulary, or a vocabulary constructed by another method.

Let $x=(b_0,\ldots,b_{n-1})$ denote the UTF-8 byte sequence of an input text, and let $\mathcal{V}$ be a vocabulary of nonempty byte strings. We assume that $\mathcal{V}$ includes all 256 single-byte tokens, ensuring that every input has a valid encoding. A segmentation is a sequence of positions $\pi=(i_0,\ldots,i_q)$ with $0=i_0<\cdots<i_q=n$, where each substring $x[i_{r-1}:i_r]$ belongs to $\mathcal{V}$. Writing $\Pi_{\mathcal{V}}(x)$ for the set of such segmentations and $q(\pi)$ for the number of tokens in $\pi$, MCE solves
\begin{equation}
    \pi^{\star}
    = \operatorname*{arg\,min}_{\pi\in\Pi_{\mathcal{V}}(x)}
    \sum_{r=1}^{q(\pi)} c(i_{r-1},i_r),
    \label{eq:mce-objective}
\end{equation}
where $c(i,j)$ denotes the cost of encoding the substring $x[i:j]$ as a single token. We define $c(i,j)$ as the sum of a unit token cost, a boundary-dependent penalty, and an optional small token preference cost.

To define the boundary penalty, we annotate the input with word-like units. A unit is a maximal run of alphabetic characters or Unicode combining marks (e.g., the acute accent in \'{e}); each remaining character, including spaces, punctuation, digits, and characters from non-alphabetic writing systems (e.g., Chinese characters), forms a singleton unit. These annotations guide candidate costs while allowing tokens to span multiple units. For byte position $i$, let $w_i$ be its unit index. For a candidate $t=x[i:j]$, let $u_{ij}=w_{j-1}-w_i+1$ denote the number of units it touches, and let $a_{ij}=1$ if byte $i$ is the first byte of unit $w_i$ and byte $j-1$ is the last byte of unit $w_{j-1}$, and $a_{ij}=0$ otherwise. The boundary penalty is defined as
\begin{equation}
    p(i,j)=
    \begin{cases}
        0, & u_{ij}=1 \text{ or } a_{ij}=1,\\
        \lambda, & u_{ij}=2 \text{ and } a_{ij}=0,\\
        +\infty, & u_{ij}\geq 3 \text{ and } a_{ij}=0,
    \end{cases}
    \label{eq:mce-boundary}
\end{equation}
where we set $\lambda=0.05$ in practice. Thus, tokens contained within one unit and tokens covering a sequence of complete units receive no boundary penalty. A token crossing one unit boundary with an incomplete endpoint incurs a small penalty, whereas a token touching three or more units with an incomplete endpoint is excluded. For example, in \texttt{co-advisor}, the units are \texttt{co}, \texttt{-}, and \texttt{advisor}. If present in the vocabulary, \texttt{co-advisor} receives no boundary penalty, \texttt{o-} incurs $\lambda$, and \texttt{o-ad} is excluded. This rule permits tokens spanning multiple complete words (e.g., phrases), potentially improving compression while discouraging partial-word boundary crossings that may yield linguistically unmotivated tokens.

Optionally, we introduce a small token-dependent cost $\varepsilon(t)\ll 1$ to express a weak preference for certain vocabulary entries. In our implementation, tokens with smaller IDs receive lower costs, using token IDs as a rough proxy for frequency to introduce a weak preference for more frequent tokens.

Combining these terms, the encoding cost of a candidate token $t=x[i:j]$ is
\begin{equation}
    c(i,j)=1+p(i,j)+\varepsilon(t),
    \label{eq:mce-cost}
\end{equation}
where the constant $1$ represents a unit cost per token. MCE minimizes this combined cost according to Equation~\ref{eq:mce-objective}, encouraging shorter token sequences while balancing token count, boundary penalties, and the optional token preference.

We solve Equation~\ref{eq:mce-objective} by dynamic programming over byte positions. Algorithm~\ref{alg:mce} summarizes the procedure. Let $D(i)$ denote the minimum cost of encoding the suffix $x[i:n]$. With $D(n)=0$, the recurrence is
\begin{equation}
    D(i)=\min_{\substack{i<j\leq n\\x[i:j]\in\mathcal{V}}}
    \left\{c(i,j)+D(j)\right\},
    \qquad i=n-1,\ldots,0.
    \label{eq:mce-dp}
\end{equation}
Storing the minimizing choice at each position allows us to recover an optimal segmentation from position zero. Since token costs are additive and independent of the segmentation of the remaining text, this recurrence yields a globally optimal segmentation.

Within CNF, MCE is used both to measure the realized token usage for vocabulary filtering and to tokenize text with the final vocabulary. More generally, MCE can operate on any vocabulary satisfying the byte-coverage assumption above, applying the same encoding procedure to produce token sequences for the language model.

\begin{algorithm}[t]
\caption{Min-Cost Encoding (MCE)}
\label{alg:mce}
\begin{algorithmic}[1]
\Require UTF-8 byte sequence $x$ of length $n$; vocabulary $\mathcal{V}$ containing all single-byte tokens; candidate costs $c(i,j)$ from Equation~\ref{eq:mce-cost}
\Ensure An optimal token sequence $T$
\State $D[n] \gets 0$
\For{$i=n-1,\ldots,0$}
    \State $\mathcal{J}_i \gets \{j \mid i<j\leq n,\ x[i:j]\in\mathcal{V}\}$
    \State $j^\star \gets \operatorname*{arg\,min}_{j\in\mathcal{J}_i}\{c(i,j)+D[j]\}$
        \Comment{Choose the largest $j$ in a tie}
    \State $D[i] \gets c(i,j^\star)+D[j^\star]$
    \State $\mathrm{next}[i] \gets j^\star$
\EndFor
\State $T \gets [\,]$, $i \gets 0$
\While{$i<n$}
    \State $j \gets \mathrm{next}[i]$
    \State Append $x[i:j]$ to $T$
    \State $i \gets j$
\EndWhile
\State \Return $T$
\end{algorithmic}
\end{algorithm}

%% file: sections/4_experiments_n.tex
\section{Experiments}
\label{sec:exp}

We evaluate CNF--MCE in terms of compression performance, vocabulary scaling, intrinsic tokenizer properties, segmentation under fixed vocabularies, and downstream language model performance.

\subsection{Experimental Setup}
\label{sec:exp-setup}

\paragraph{Data.}
Our tokenizer experiments cover six text categories: English web text (\texttt{en-web}), English PDF text (\texttt{en-pdf}), mathematics (\texttt{math}), code (\texttt{code}), Chinese text (\texttt{zh}), and multilingual text (\texttt{lang11}). The code and multilingual data span 14 programming languages and 11 languages, respectively. The datasets and their corresponding categories are listed in Appendix~\ref{app:tokenizer-data}.

\paragraph{Tokenizer configuration.}
We construct our CNF vocabularies using a mixture of 6.9 million documents sampled from these data sources, comprising English (1M), mathematics (0.2M), code (1.7M), Chinese text (1M), and multilingual text (3M). For CNF, we use three counting rounds with candidate length intervals $R=([1,12],[13,21],[22,30])$, corresponding frequency thresholds $t_r$ of $0$, $80$, and $80$, respectively, and a raw vocabulary size of $m=5{,}000{,}000$. We set the target final vocabulary size to $k=138{,}000$ and $k=250{,}000$ to construct \texttt{mix138k} and \texttt{mix250k}, respectively. We encode text with the resulting vocabularies using MCE, with the boundary penalty coefficient $\lambda=0.05$ and the token preference cost $\varepsilon(t)=0.1/(10+s(t))$, where $s(t)=10{,}000/(1+0.01\cdot\operatorname{id}(t))$ and $\operatorname{id}(t)$ denotes the ID of token $t$ in the vocabulary. This cost assigns a smaller penalty to tokens with smaller IDs.

\subsection{Compression Performance across Tokenizers}
\label{sec:exp-compression}

We compare the compression performance of CNF--MCE with existing BPE tokenizers used by LLMs. The baselines use their corresponding BPE encoding procedures, while our Mix vocabularies are constructed by CNF and encoded with MCE. We consider two vocabulary-size groups: 100k--160k entries, including our \texttt{mix138k} vocabulary, and 200k--250k entries, including our \texttt{mix250k} vocabulary. Table~\ref{tab:vocab-list} summarizes the vocabularies used in our experiments.

\begin{table}[htbp]
    \centering
    \caption{Overview of the vocabularies used in our experiments.}
    \label{tab:vocab-list}
    \footnotesize
    \setlength{\tabcolsep}{1.5pt}
    \begin{tabular}{llp{0.13\linewidth}rp{0.53\linewidth}}
        \toprule
        Group & Vocab. name & Construction & Size & Model series \\
        \midrule
        \multirow{5}{*}{\makecell{Smaller\\(100k--160k)}} & llama128k    & BPE & 128k & Llama3.2 \citep{meta2024llama32} \\
                                & ds128k       & BPE & 128k & DeepSeek-V4 \citep{deepseekai2026deepseekv4} \\
                                & qwen152k     & BPE & 152k & Qwen2.5/3 \citep{qwen2025qwen25technicalreport,qwen3technicalreport} \\
                                & cl100k\_base & BPE & 100k & GPT-3.5/4 \citep{openai2026tiktoken} \\
                                & mix138k      & CNF & 138k & -- \\
        \midrule
        \multirow{4}{*}{\makecell{Larger\\(200k--250k)}}  & o200k\_base  & BPE & 200k & GPT-4o/4.1/4.5, o1, o3, o4 \citep{openai2026tiktoken} \\
                                & qwen250k     & BPE & 248k & Qwen3.5/3.6/3.8 \citep{qwen2026qwen35,qwen2026qwen36,qwen2026qwen38} \\
                                & mm200k       & BPE & 200k & MiniMax-M2.5/M3 \citep{chen2026minimaxm2} \\
                                & mix250k      & CNF & 249k & -- \\
        \bottomrule
    \end{tabular}%
\end{table}

\paragraph{Evaluation protocol.}
We evaluate on six data categories: English web text (\texttt{en-web}), English PDF text (\texttt{en-pdf}), mathematics (\texttt{math}), code (\texttt{code}), Chinese text (\texttt{zh}), and multilingual text (\texttt{lang11}). For each category, we randomly sample 10,000 documents and evaluate all tokenizers on the same sampled documents. We measure the character-level compression rate in characters per token (CPT):
\begin{equation}
    \mathrm{CPT}(T,\mathcal{D})
    = \frac{\sum_{x\in\mathcal{D}} |x|_{\mathrm{char}}}
           {\sum_{x\in\mathcal{D}} |T(x)|},
    \label{eq:exp-cpt}
\end{equation}
where $\mathcal{D}$ denotes the sampled documents, $|x|_{\mathrm{char}}$ is the number of characters in document $x$, and $|T(x)|$ is the number of tokens produced by tokenizer $T$. Higher CPT indicates that fewer tokens are required to encode the same text.

\begin{table}[t]
    \centering
    \caption{Compression performance in the Smaller-vocabulary group.}
    \label{tab:token-efficiency-100k}
    \resizebox{\linewidth}{!}{%
    \begin{tabular}{lrrrrrrr}
        \toprule
        \multirow{2}{*}{Text type} & \multirow{2}{*}{llama128k} & \multirow{2}{*}{ds128k} & \multirow{2}{*}{qwen152k} & \multirow{2}{*}{cl100k\_base} & \multirow{2}{*}{mix138k} & \multicolumn{2}{c}{Improvement over} \\
        \cmidrule(lr){7-8}
        & & & & & & ds128k & qwen152k \\
        \midrule
        en-web & 4.640 & 4.643 & 4.545 & 4.632 & \textbf{5.599} & 20.6\% & 23.2\% \\
        en-pdf & 4.023 & 4.145 & 3.905 & 4.013 & \textbf{4.936} & 19.1\% & 26.4\% \\
        math   & 3.640 & 3.725 & 3.471 & 3.630 & \textbf{4.462} & 19.8\% & 28.6\% \\
        code   & 3.413 & 3.178 & 3.143 & 3.381 & \textbf{3.728} & 17.3\% & 18.6\% \\
        zh     & 1.240 & 1.696 & 1.553 & 0.872 & \textbf{1.812} & 6.8\%  & 16.7\% \\
        lang11 & 3.135 & 3.213 & 3.177 & 2.771 & \textbf{4.028} & 25.4\% & 26.8\% \\
        \bottomrule
    \end{tabular}%
    }
\end{table}

\begin{table}[t]
    \centering
    \caption{Compression performance in the Larger-vocabulary group.}
    \label{tab:token-efficiency-200k}
    \resizebox{\linewidth}{!}{%
    \begin{tabular}{lrrrrrr}
        \toprule
        \multirow{2}{*}{Text type} & \multirow{2}{*}{o200k\_base} & \multirow{2}{*}{qwen250k} & \multirow{2}{*}{mm200k} & \multirow{2}{*}{mix250k} & \multicolumn{2}{c}{Improvement over} \\
        \cmidrule(lr){6-7}
        & & & & & o200k\_base & qwen250k \\
        \midrule
        en-web & 4.700 & 4.499 & 4.752 & \textbf{5.955} & 26.7\% & 32.4\% \\
        en-pdf & 4.051 & 3.856 & 4.244 & \textbf{5.211} & 28.7\% & 35.1\% \\
        math   & 3.651 & 3.400 & 3.774 & \textbf{4.678} & 28.1\% & 37.6\% \\
        code   & 3.388 & 2.983 & 3.324 & \textbf{3.880} & 14.5\% & 30.1\% \\
        zh     & 1.311 & 1.703 & 1.758 & \textbf{1.963} & 49.7\% & 15.3\% \\
        lang11 & 3.648 & 3.748 & 3.434 & \textbf{4.530} & 24.2\% & 20.9\% \\
        \bottomrule
    \end{tabular}%
    }
\end{table}

\paragraph{Results with Smaller vocabularies.}
Table~\ref{tab:token-efficiency-100k} compares \texttt{mix138k} with \texttt{llama128k}, \texttt{ds128k}, \texttt{qwen152k}, and \texttt{cl100k\_base}. Our tokenizer achieves the highest CPT in all six categories. On English web text, it achieves 5.599 characters per token, improving over \texttt{ds128k} and \texttt{qwen152k} by 20.6\% and 23.2\%, respectively. The gains also extend to multilingual text, where CPT increases by 25.4\% and 26.8\% over these two baselines.

\paragraph{Results with Larger vocabularies.}
Table~\ref{tab:token-efficiency-200k} compares \texttt{mix250k} with \texttt{o200k\_base}, \texttt{qwen250k}, and \texttt{mm200k}. Our tokenizer again achieves the highest CPT across all categories. Relative to \texttt{qwen250k}, improvements range from 15.3\% on Chinese text to 37.6\% on mathematics. Relative to \texttt{o200k\_base}, the largest improvement is observed on Chinese text, with a gain of 49.7\%. 

These results demonstrate consistent compression gains for CNF--MCE across all evaluated text categories and both vocabulary-size groups. These compression gains can be particularly beneficial for long-context processing. With standard full attention, attention computation scales quadratically with sequence length, so encoding the same text with fewer tokens can substantially reduce computation and potentially lower serving latency and cost. Moreover, shorter sequences reduce KV-cache storage, easing GPU memory pressure during inference.

\subsection{Vocabulary Scaling}
\label{sec:exp-vocab-scaling}

We compare BPE and CNF--MCE across vocabulary sizes to investigate whether token efficiency remains scalable as the vocabulary grows larger.

\paragraph{Evaluation protocol.}
We evaluate on the English web text (\texttt{en-web}), with vocabulary sizes doubling from 1K to 1M. For each vocabulary and encoding method, we process documents until the cumulative output reaches 100M tokens. We use a fixed token budget to ensure a fair comparison of vocabulary utilization across configurations. Encoding efficiency is measured in characters per token (CPT), as in Equation~\ref{eq:exp-cpt}, over the documents processed by each configuration. The \emph{vocabulary utilization} is defined as the fraction of vocabulary entries observed at least once: $U(\mathcal{V}) = |\{t \in \mathcal{V} : f(t) > 0\}| / |\mathcal{V}|$, where $f(t)$ is the observed frequency of token $t$ in the generated token sequence. Higher values indicate broader vocabulary usage. Detailed results are provided in Appendix~\ref{app:vocab-scaling}.

\begin{figure}[t]
    \centering
    \includegraphics[width=\linewidth]{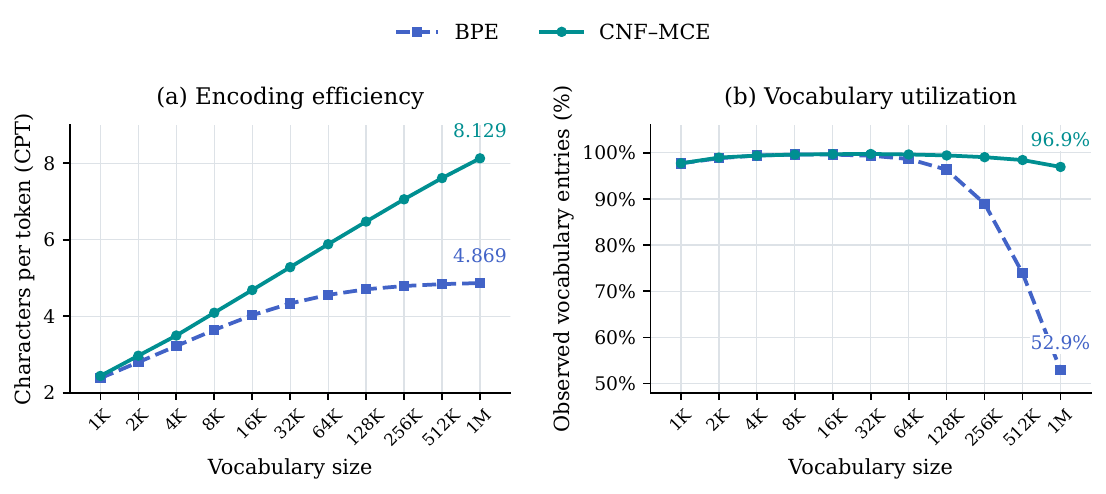}
    \caption{Vocabulary scaling on \texttt{en-web}: (a) token efficiency and (b) vocabulary utilization after 100M output tokens per configuration. Higher is better in both panels. Vocabulary size uses a logarithmic axis; 1K denotes 1,000 entries and 1M denotes 1,024,000 entries.}
    \label{fig:vocab-scaling}
\end{figure}

\paragraph{Token efficiency.}
Figure~\ref{fig:vocab-scaling}(a) shows that CNF--MCE achieves higher token efficiency than BPE across all evaluated vocabulary sizes. BPE's token efficiency begins to saturate as the vocabulary grows, whereas CNF--MCE exhibits stronger scaling with sustained gains. CNF--MCE continues to improve even when the vocabulary doubles from 512K to 1M, gaining 6.8\% in CPT compared with just 0.6\% for BPE. At 1M entries, CNF--MCE achieves 67.0\% higher CPT than BPE.

\paragraph{Vocabulary utilization.}
Figure~\ref{fig:vocab-scaling}(b) shows that BPE's token efficiency saturation is accompanied by a marked decline in vocabulary utilization as the vocabulary grows. In contrast, CNF--MCE maintains high utilization even at 1M entries, reaching 96.9\% compared with 52.9\% for BPE under the same token budget.

These results indicate that CNF--MCE exhibits stronger vocabulary scaling than BPE, with sustained gains in token efficiency and high vocabulary utilization up to 1M entries.

\subsection{Intrinsic Evaluation}
\label{sec:exp-intrinsic}

We further compare \texttt{qwen250k} and \texttt{mix250k} across a broader set of intrinsic metrics, which characterize task-independent tokenizer properties.

\paragraph{Evaluation protocol.}
We evaluate both tokenizers on a subset of FLORES-200 \citep{nllb2022} covering 13 languages using the third-party evaluation toolkit TokEval \citep{meister2026tokeval}, with the metrics and results summarized in Table~\ref{tab:intrinsic-tokeval}. \emph{Compression rate} is measured at the byte level as the number of UTF-8 bytes per token (bytes/token). \emph{Fertility} and \emph{token length} characterize segmentation granularity, while \emph{vocabulary utilization} measures the fraction of vocabulary entries used in the evaluation corpus. \emph{R\'enyi efficiency} characterizes the token usage distribution, while \emph{bigram entropy} and \emph{trigram entropy} measure normalized conditional entropy given the preceding one and two tokens, respectively. We assess cross-lingual encoding inequality using \emph{tokenizer fairness Gini}, and numerical tokenization using \emph{three-digit boundary alignment}, \emph{numeric magnitude consistency}, and \emph{operator isolation rate}. Finally, \emph{reconstruction fidelity} and \emph{UTF-8 token integrity} measure exact text reconstruction and token completeness, respectively. We retain the metric conventions and directions shown in the evaluation report.

\begin{table}[t]
    \centering
    \caption{Intrinsic evaluation with TokEval.}
    \label{tab:intrinsic-tokeval}
    \small
    \begin{tabular}{lrr}
        \toprule
        Metric & qwen250k & mix250k \\
        \midrule
        \multicolumn{3}{l}{\textit{Compression and vocabulary usage}} \\
        Fertility $\downarrow$                  & 5.550  & \textbf{4.429} \\
        Token length $\uparrow$                 & 3.832  & \textbf{4.718} \\
        Vocabulary utilization $\uparrow$       & 0.2147 & \textbf{0.2696} \\
        Compression rate (bytes/token) $\uparrow$ & 5.143  & \textbf{6.329} \\
        \midrule
        \multicolumn{3}{l}{\textit{Token distributions and multilingual fairness}} \\
        R\'enyi efficiency ($\alpha=2$) $\uparrow$ & 0.4948 & \textbf{0.5476} \\
        Bigram entropy $\uparrow$               & 0.8805 & \textbf{0.9438} \\
        Trigram entropy $\uparrow$              & \textbf{0.8151} & 0.7746 \\
        Tokenizer fairness Gini $\downarrow$    & 0.1214 & \textbf{0.1176} \\
        \midrule
        \multicolumn{3}{l}{\textit{Numerical tokenization}} \\
        Three-digit boundary alignment (F1) $\uparrow$ & 0.3703 & \textbf{0.7655} \\
        Numeric magnitude consistency $\downarrow$ & 1.0000 & \textbf{0.5633} \\
        Operator isolation rate $\uparrow$      & 0.4863 & \textbf{0.4890} \\
        \midrule
        \multicolumn{3}{l}{\textit{Encoding fidelity}} \\
        Reconstruction fidelity (exact match) $\uparrow$ & \textbf{1.0000} & \textbf{1.0000} \\
        UTF-8 token integrity $\uparrow$        & 0.9994 & \textbf{0.9996} \\
        \bottomrule
    \end{tabular}
\end{table}

\paragraph{Results.}
\texttt{mix250k} achieves a compression rate of 6.329 compared with 5.143 for \texttt{qwen250k}, an improvement of approximately 23.1\%. Fertility decreases by 20.2\%, while mean token length increases by 23.1\%. Vocabulary utilization rises by 25.6\%, indicating that a larger fraction of the vocabulary is used in the evaluation corpus. These results show that \texttt{mix250k} achieves more compact tokenization alongside broader vocabulary usage.

R\'enyi efficiency increases by 10.7\%, indicating more evenly distributed token usage. Bigram entropy also increases, while trigram entropy decreases. The lower tokenizer fairness Gini suggests a modest reduction in cross-lingual encoding inequality on the evaluation corpus.

The numerical tokenization results show consistent improvements across all three metrics, with the largest relative gain of 106.7\% in three-digit boundary alignment F1, suggesting that \texttt{mix250k} handles numerical tokenization more effectively than \texttt{qwen250k}. Both tokenizers achieve an exact reconstruction rate of 1.0, while \texttt{mix250k} has slightly higher UTF-8 token integrity.

Overall, \texttt{mix250k} outperforms \texttt{qwen250k} on most reported intrinsic metrics. It offers better compression and higher vocabulary utilization.

\subsection{Segmentation with Fixed Vocabularies}
\label{sec:exp-fixed-vocab}

We compare various segmentation algorithms under fixed vocabularies to examine how the encoding algorithm affects compression and segmentation.

\paragraph{Evaluation protocol.}
We compare BPE, MCE, and forward maximum matching (FMM) on the BPE-built \texttt{qwen250k} vocabulary, and MCE and FMM on the CNF-built \texttt{mix250k} vocabulary.
Note that BPE tokenizaiton cannot be applied to mix250k, as no merge information is available in this vocabulary.
FMM segments text from left to right by selecting the longest matching vocabulary entry at each position.

We evaluate \emph{token efficiency} using the number of tokens per 1,000 characters (tokens/kchar). Lower values indicate that fewer tokens are needed to encode the same text, corresponding to better compression. We further measure the similarity between two segmentations using two metrics. \emph{Distance} (Dist.) is the edit distance between the token-ID sequences produced by two algorithms, divided by the length of the longer sequence. \emph{Agreement} (Agree.) measures consistency at the byte level: each byte in the original text is assigned the ID of the token covering it, and agreement is the proportion of byte positions assigned the same token ID by both algorithms. Lower Distance and higher Agreement indicate more similar segmentations. We also report \emph{Boundary-crossing}, the proportion of tokens in a segmentation that incompletely span multiple word-like units, where the units are defined in
Section~\ref{sec:mce}. A token counts as a crossing if it spans multiple units but starts inside the first unit or ends inside the last, as the token ``\texttt{is\textvisiblespace a}'' in ``\texttt{is\textvisiblespace ancient}''. Tokens covering a sequence of complete units do not count as crossings.

\begin{table}[t]
    \centering
    \caption{Token efficiency and pairwise segmentation comparisons on the fixed \texttt{qwen250k} vocabulary.}
    \label{tab:segmentation-qwen}
    \setlength{\tabcolsep}{4pt}
    \resizebox{\linewidth}{!}{%
    \begin{tabular}{lrrrrrrrrr}
        \toprule
        \multirow{2}{*}{Text type} & \multicolumn{3}{c}{Token efficiency} & \multicolumn{2}{c}{MCE vs. BPE} & \multicolumn{2}{c}{FMM vs. BPE} & \multicolumn{2}{c}{FMM vs. MCE} \\
        \cmidrule(lr){2-4}\cmidrule(lr){5-6}\cmidrule(lr){7-8}\cmidrule(lr){9-10}
        & BPE & FMM & MCE & Dist. (\%) & Agree. (\%) & Dist. (\%) & Agree. (\%) & Dist. (\%) & Agree. (\%) \\
        \midrule
        en-web      & 222.5 & 222.1 & \textbf{221.6} & 3.64  & 97.53 & 7.30 & 94.78 & 5.95 & 95.60 \\
        math        & 297.6 & 297.2 & \textbf{296.6} & 3.61  & 96.88 & 6.54 & 94.02 & 5.52 & 94.95 \\
        code/python & 278.3 & 278.0 & \textbf{276.4} & 11.21 & 87.50 & 21.48 & 70.91 & 16.4 & 78.26 \\
        zh      & 643.7 & 643.4 & \textbf{637.4} & 2.65  & 97.00 & 6.75 & 92.11 & 5.21 & 93.72 \\
        \bottomrule
    \end{tabular}%
    }
\end{table}

\begin{table}[t]
    \centering
    \caption{Compression and segmentation comparisons on the fixed \texttt{mix250k} vocabulary.}
    \label{tab:segmentation-mix}
    \resizebox{\linewidth}{!}{%
    \begin{tabular}{lrrrrrr}
        \toprule
        \multirow{2}{*}{Text type} & \multicolumn{2}{c}{Token efficiency} & \multicolumn{2}{c}{MCE vs. FMM} & \multicolumn{2}{c}{Boundary-crossing (\%)} \\
        \cmidrule(lr){2-3}\cmidrule(lr){4-5}\cmidrule(lr){6-7}
        & FMM & MCE & Dist. (\%) & Agree. (\%) & MCE & FMM \\
        \midrule
        en-web        & 170.4 & \textbf{168.0} & 20.73 & 83.10 & 0 & 2.291 \\
        math/finemath & 223.6 & \textbf{221.5} & 17.23 & 84.31 & 0 & 1.210 \\
        code/python   & 234.3 & \textbf{230.1} & 23.3  & 78.23 & 0 & 1.064 \\
        zh        & 510.6 & \textbf{509.3} & 11.62 & 88.32 & 0 & 0.019 \\
        \bottomrule
    \end{tabular}%
    }
\end{table}

\paragraph{Results with the BPE vocabulary.}
Table~\ref{tab:segmentation-qwen} compares the three encoding algorithms on \texttt{qwen250k}. MCE produces the fewest tokens across all four datasets.
These results show that MCE can improve compression even when using a vocabulary constructed by BPE.

The table also reports pairwise segmentation comparisons. \emph{Agreement} between MCE and BPE exceeds 96\% on English web, mathematics, and Chinese text. Thus, without relying on BPE's merge order during encoding, MCE produces segmentations highly similar to those of BPE while improving compression. This demonstrates its applicability to the existing BPE-built vocabulary. Agreement is lower on Python code (87.50\%), indicating that the degree of similarity varies across domains.

Two examples illustrate how the algorithms differ in segmentation. For text ``\texttt{(taxonomy}'' (not including the quotes), BPE produces ``\texttt{(t$\|$ax$\|$onomy}'' and FMM produces ``\texttt{(t$\|$axon$\|$omy}'', whereas MCE yields ``\texttt{($\|$taxonomy}'', reducing the token count from three to two while keeping the word intact. For text ``\texttt{memory\_efficient}'', all three methods produce three tokens: BPE and FMM yield ``\texttt{memory$\|$\_eff$\|$icient}'', while MCE yields ``\texttt{memory$\|$\_$\|$efficient}'', separating the underscore and preserving both words as complete tokens. More examples are provided in Table~\ref{tab:tokenization-examples}.

\paragraph{Results with the CNF vocabulary.}
Table~\ref{tab:segmentation-mix} compares MCE and FMM on \texttt{mix250k}. MCE consistently achieves greater token efficiency than FMM across all four data categories. The efficiency advantage is more pronounced when comparing MCE with \texttt{mix250k} against BPE with \texttt{qwen250k} (e.g., on English web text, token usage decreases from 222.5 to 168.0 tokens/kchar, a reduction of approximately 24.5\%). Pairwise comparisons reveal noticeable differences between the segmentations produced by MCE and FMM under the same vocabulary. We further examine Boundary-crossing: MCE exhibits no incomplete crossings of multiple word-like units across the evaluated data categories, whereas FMM has nonzero rates in all four. These results suggest that MCE improves token efficiency while avoiding partial-word boundary crossings, thereby helping to preserve the semantic integrity of the resulting tokens.

These results demonstrate the effectiveness of MCE across vocabularies. MCE keeps high token efficiency on both BPE- and CNF-built vocabularies without relying on merge rules, while avoiding incomplete crossings of word-like units.

\subsection{Downstream Task Evaluation}
\label{sec:exp-downstream}

Although our method has demonstrated advantages in intrinsic tokenizer properties, further evaluation on downstream tasks is needed to assess its practical effectiveness. We train language models from scratch that differ only in their tokenizers and evaluate them across a range of benchmarks.

\paragraph{Evaluation protocol.}
We train language models from scratch at the 1.8B- and 8B-parameter scales. At 1.8B, we compare tokenizers using the BPE-built \texttt{qwen152k} vocabulary and our CNF-built \texttt{mix138k} vocabulary; at 8B, we compare \texttt{qwen152k} with our CNF-built \texttt{mix250k} vocabulary. Within each model scale, only the tokenizer differs; the remaining model and training settings are held fixed. Our training recipe is based on SmolLM3 \citep{bakouch2025smollm3}. We evaluate checkpoints at the end of the \emph{phase-1} and the \emph{decay phase}. Detailed model configurations are provided in Appendix~\ref{app:model-configurations}.

We evaluate on 11 English benchmarks, described in Appendix~\ref{app:downstream-benchmarks}. These benchmarks cover five categories: knowledge (ARC, MMLU, OpenBookQA, and TriviaQA), math (GSM8K), coding (HumanEval), commonsense (CommonsenseQA, PIQA, and WinoGrande), and language understanding (HellaSwag and LAMBADA). All tasks use zero-shot evaluation except GSM8K, which uses five-shot evaluation. We report exact match (EM) for GSM8K, LAMBADA, and TriviaQA, pass@1 for HumanEval, and accuracy under cloze prompting (CP) for the remaining tasks. The average is the unweighted mean across the 11 benchmarks. Further evaluation details are provided in Appendix~\ref{app:downstream-metrics}.

\begin{table}[t]
    \centering
    \caption{Downstream performance of language models trained from scratch at the 1.8B scale with different vocabularies.}
    \label{tab:downstream}
    \setlength{\tabcolsep}{6pt}
    \small
    \begin{tabular}{lrrrr}
        \toprule
        \multirow{2}{*}{Benchmark} & \multicolumn{2}{c}{Phase-1} & \multicolumn{2}{c}{Decay phase} \\
        \cmidrule(lr){2-3}\cmidrule(lr){4-5}
        & \makecell{qwen152k} & \makecell{mix138k} & \makecell{qwen152k} & \makecell{mix138k} \\
        \midrule
        ARC          & \textbf{53.2} & 49.4          & 58.1          & \textbf{58.2} \\
        CommonsenseQA & \textbf{39.6} & 38.7          & \textbf{38.5} & 36.1          \\
        GSM8K        & 0.7           & \textbf{2.1}  & 17.9          & \textbf{20.6} \\
        HellaSwag    & 59.2          & \textbf{59.9} & 61.2          & \textbf{61.7} \\
        HumanEval    & 1.0           & \textbf{1.7}  & \textbf{6.0}  & 4.3           \\
        LAMBADA      & 56.3          & \textbf{59.9} & 58.1          & \textbf{61.0} \\
        MMLU         & \textbf{36.8} & 34.3          & \textbf{40.7} & 38.0          \\
        OpenBookQA   & 39.3          & \textbf{43.2} & 40.9          & \textbf{43.0} \\
        PIQA         & \textbf{74.0} & 73.3          & \textbf{74.6} & 73.8          \\
        TriviaQA     & \textbf{28.0} & 26.2          & \textbf{31.7} & 30.7          \\
        WinoGrande   & \textbf{57.0} & 55.7          & 57.1          & \textbf{58.2} \\
        \midrule
        Average & \textbf{40.5} & 40.4          & \textbf{44.1} & \textbf{44.1} \\
        \bottomrule
    \end{tabular}%
\end{table}

\begin{table}[t]
    \centering
    \caption{Downstream performance of language models trained from scratch at the 8B scale with different vocabularies.}
    \label{tab:downstream-8b}
    \setlength{\tabcolsep}{6pt}
    \small
    \begin{tabular}{lrrrr}
        \toprule
        \multirow{2}{*}{Benchmark} & \multicolumn{2}{c}{Phase-1} & \multicolumn{2}{c}{Decay phase} \\
        \cmidrule(lr){2-3}\cmidrule(lr){4-5}
        & \makecell{qwen152k} & \makecell{mix250k} & \makecell{qwen152k} & \makecell{mix250k} \\
        \midrule
        ARC            & \textbf{59.9} & 57.1 & \textbf{65.0} & 62.7 \\
        CommonsenseQA  & \textbf{41.8} & 41.6 & 41.4 & \textbf{42.9} \\
        GSM8K          & 1.2 & \textbf{6.1} & 20.8 & \textbf{38.7} \\
        HellaSwag      & 67.2 & \textbf{67.6} & \textbf{72.1} & 70.8 \\
        HumanEval      & \textbf{1.8} & 1.2 & \textbf{11.6} & 4.3 \\
        LAMBADA        & 57.8 & \textbf{68.7} & 64.4 & \textbf{72.2} \\
        MMLU           & \textbf{41.5} & 38.5 & \textbf{46.1} & 42.2 \\
        OpenBookQA     & 42.3 & \textbf{45.2} & 43.2 & \textbf{46.0} \\
        PIQA           & 77.4 & \textbf{77.7} & 79.1 & \textbf{79.2} \\
        TriviaQA       & \textbf{44.7} & 41.8 & \textbf{53.0} & 48.8 \\
        WinoGrande     & \textbf{61.9} & 59.1 & \textbf{64.2} & 59.5 \\
        \midrule
        Average & 45.2 & \textbf{45.9} & 51.0 & \textbf{51.6} \\
        \bottomrule
    \end{tabular}%
\end{table}

\paragraph{Results at the 1.8B scale.}
Table~\ref{tab:downstream} shows that \texttt{mix138k} maintains comparable average downstream performance to \texttt{qwen152k} at the 1.8B scale. The phase-1 average scores are 40.4 and 40.5, respectively. Both improve after the decay phase, with average scores rounding to 44.1 for both configurations. Most tasks improve over this transition: 9 of 11 for \texttt{mix138k} and 10 of 11 for \texttt{qwen152k}.

Across individual tasks, \texttt{mix138k} remains broadly competitive with \texttt{qwen152k}, scoring higher on HellaSwag, LAMBADA, and OpenBookQA in both phases. After the decay phase, it outperforms \texttt{qwen152k} on 6 of the 11 benchmarks: ARC, GSM8K, HellaSwag, LAMBADA, OpenBookQA, and WinoGrande. The largest advantages are on LAMBADA, GSM8K, and OpenBookQA, with gains of 2.9, 2.7, and 2.1 points, respectively. These gains are accompanied by lower scores on MMLU, CommonsenseQA, and HumanEval by 2.7, 2.4, and 1.7 points, respectively. Thus, the compression advantage of \texttt{mix138k} is accompanied by comparable average downstream performance, with task-specific trade-offs.

\paragraph{Results at the 8B scale.}
Table~\ref{tab:downstream-8b} compares \texttt{mix250k} with \texttt{qwen152k} at the 8B scale. The 8B models further improve performance over the 1.8B models. The two configurations achieve similar average downstream performance, with \texttt{mix250k} scoring slightly higher in both the phase-1 (45.9 vs. 45.2) and the decay phase (51.6 vs. 51.0). From the phase-1 to the  decay phase, all 11 tasks improve for \texttt{mix250k}, compared  with 10 of 11 for \texttt{qwen152k}. \texttt{mix250k} scores higher on 5 of the 11 benchmarks, with the largest gains on GSM8K (+17.9 points) and LAMBADA (+7.8 points) at decay phase.

We use \texttt{qwen152k} as a common baseline across both model scales. The 8B experiment assesses the downstream effectiveness of our larger \texttt{mix250k} vocabulary relative to this baseline. Since the vocabularies differ in size, we leave a downstream comparison between \texttt{qwen250k} and \texttt{mix250k} to
future work to provide a more controlled evaluation at comparable vocabulary sizes.

Overall, our method achieves better compression while maintaining competitive downstream performance at both model scales.

%% file: sections/5_conclusion.tex
\section{Conclusion}
\label{sec:conclu}

We introduced CNF--MCE, a merge-free, non-iterative tokenization framework that separates vocabulary construction from segmentation. CNF combines direct substring counting with a single filtering step based on realized token usage. MCE globally minimizes an encoding cost combining token count with penalties for partial-word boundary crossings while allowing multiword tokens. CNF--MCE offers several advantages over BPE, including higher token efficiency, greater scalability, and lower dependency. Experiments show consistent compression gains over the evaluated BPE tokenizers across six text categories and improvements in other intrinsic metrics. On English web text, CNF--MCE sustains compression gains and high vocabulary utilization up to 1M entries. Models trained at the 1.8B and 8B scales achieve comparable average performance to the Qwen tokenizer baseline across 11 English benchmarks. These findings support CNF-MCE as an effective approach to improving token efficiency while preserving competitive downstream performance. Future work will involve comparing downstream performance across a wider range of vocabulary sizes, model architectures, model sizes, languages, and benchmarks.

%% file: sections/appendix.tex
\newpage
\section{Tokenizer Details}

\subsection{Data Sources for Tokenizer Training and Testing}
\label{app:tokenizer-data}

Table~\ref{tab:tokenizer-data-sources} lists the data sources used in our tokenizer experiments. We distinguish English web text (\texttt{en-web}) from English PDF text (\texttt{en-pdf}) for evaluation. The code data cover 14 programming languages, and the multilingual data (\texttt{lang11}) cover 11 human languages.
Please note that most of the listed datasets are quite large, and only a small portion of them are used for experiments. When a dataset is used for both training and testing, different randomly sampled subsets are used for each purpose. 

\begin{table}[t]
    \centering
    \caption{Data sources used in the tokenizer experiments. Dolma3 subsets refer to the \texttt{dolma3\_mix-5.5T-1125} release.}
    \label{tab:tokenizer-data-sources}
    \small
    \begin{tabular}{@{}lp{0.7\linewidth}l@{}}
        \toprule
        Text type & Datasets and subsets & Used for \\
        \midrule
        en-web & FineWeb-Edu \citep{lozhkov2024fineweb-edu}; DCLM-Edu \citep{allal2025smollm2smolgoesbig}; the \path{common_crawl} subset of Dolma3 \citep{olmo2025olmo3}. & train \& test \\
        \addlinespace
        en-pdf & The \path{olmocr_science_pdfs} and \path{rpj-proofpile-arxiv} subsets of Dolma3 \citep{olmo2025olmo3}; peS2o \citep{peS2o}. & test \\
        \addlinespace
        math & FineMath \citep{allal2025smollm2smolgoesbig}; Nemotron-CC-Math-v1 \citep{karimi2025nemotroncc}. & train \& test \\
        \addlinespace
        code & The Stack v2 deduplicated release (\path{the-stack-v2-dedup}) \citep{lozhkov2024starcoder}. & train \& test \\
        \addlinespace
        zh & FineWeb2-HQ \citep{messmer2025multilingdatacomp}; SkyPile-150B \citep{wei2023skywork}; CCI4.0 (\path{BAAI/CCI4.0-M2-Base-v1}) \citep{liu2025cci40bilingualpretrainingdataset}. & train \& test \\
        \addlinespace
        lang11 & FineWeb2 \citep{penedo2025fineweb2pipelinescale}; FineWeb2-HQ \citep{messmer2025multilingdatacomp}; Thai data from Mangosteen \citep{phatthiyaphaibun2025mangosteenopenthaicorpus} and WanJuanSiLu \citep{yu2025wanjuansiluhighqualityopensourcewebtext}. & train \& test \\
        \bottomrule
    \end{tabular}
\end{table}

\subsection{Vocabulary Scaling Results}
\label{app:vocab-scaling}

Table~\ref{tab:vocab-scaling} provides the detailed results for the vocabulary scaling experiment in Section~\ref{sec:exp-vocab-scaling}.

\begin{table}[t]
    \centering
    \caption{Vocabulary scaling on \texttt{en-web} with a 100M-token budget per configuration. CPT denotes characters per token; utilization is the percentage of vocabulary entries observed at least once. Higher is better for both metrics.}
    \label{tab:vocab-scaling}
    \small
    \begin{tabular}{lrrrr}
        \toprule
        \multirow{2}{*}{Vocab. size} & \multicolumn{2}{c}{BPE} & \multicolumn{2}{c}{CNF--MCE} \\
        \cmidrule(lr){2-3}\cmidrule(lr){4-5}
        & CPT $\uparrow$ & Utilization (\%) $\uparrow$ & CPT $\uparrow$ & Utilization (\%) $\uparrow$ \\
        \midrule
        1K & 2.382 & 97.7 & 2.441 & 97.7 \\
        2K & 2.800 & 98.8 & 2.970 & 99.0 \\
        4K & 3.223 & 99.4 & 3.497 & 99.4 \\
        8K & 3.644 & 99.6 & 4.093 & 99.6 \\
        16K & 4.028 & 99.6 & 4.687 & 99.7 \\
        32K & 4.336 & 99.4 & 5.284 & 99.7 \\
        64K & 4.559 & 98.6 & 5.884 & 99.6 \\
        128K & 4.707 & 96.3 & 6.475 & 99.4 \\
        256K & 4.792 & 89.0 & 7.058 & 99.0 \\
        512K & 4.841 & 73.9 & 7.613 & 98.4 \\
        1M & 4.869 & 52.9 & 8.129 & 96.9 \\
        \bottomrule
    \end{tabular}
\end{table}

\subsection{Tokenization Examples}
\label{app:tokenization-examples}

Table~\ref{tab:tokenization-examples} compares segmentations produced by BPE, FMM, and MCE using the fixed \texttt{qwen250k} vocabulary. Double vertical bars indicate token boundaries, and `\textvisiblespace{}' denotes a space.
Table~\ref{tab:tokenization-examples-mix250k} is the comparison results on the \texttt{mix250k} vocabulary. Please note that BPE is not applicable to mix250k, because no merge information is contained in this vocabulary.

\begin{table}[t]
    \centering
    \caption{Example tokenization results of various encoding algorithms operating on the qwen250k vocabulary}
    \label{tab:tokenization-examples}
    \small
    \begin{tabular}{llll}
        \toprule
        Text & BPE & FMM & MCE \\
        \midrule
        \textvisiblespace autoregulation & \textvisiblespace autore$\|$g$\|$ulation & \textvisiblespace autore$\|$gul$\|$ation & \textvisiblespace auto$\|$reg$\|$ulation \\
        \textvisiblespace pandemics & \textvisiblespace pand$\|$emics & \textvisiblespace pandemic$\|$s & \textvisiblespace pandemic$\|$s \\
        \textvisiblespace inconceivable & \textvisiblespace incon$\|$ce$\|$ivable & \textvisiblespace incon$\|$ce$\|$ivable & \textvisiblespace inc$\|$once$\|$ivable \\
        \textvisiblespace Modeller & \textvisiblespace Mod$\|$eller & \textvisiblespace Modelle$\|$r & \textvisiblespace Model$\|$ler \\
        \textvisiblespace Trips & \textvisiblespace Tri$\|$ps & \textvisiblespace Trip$\|$s & \textvisiblespace Tr$\|$ips \\
        \textvisiblespace mediate & \textvisiblespace med$\|$iate & \textvisiblespace media$\|$te & \textvisiblespace $\|$mediate \\
        \textvisiblespace arterioles & \textvisiblespace arter$\|$io$\|$les & \textvisiblespace arter$\|$iol$\|$es & \textvisiblespace art$\|$eri$\|$oles \\
        \textvisiblespace ejecting & \textvisiblespace eject$\|$ing & \textvisiblespace eject$\|$ing & \textvisiblespace ejec$\|$ting \\
        \textvisiblespace too-little-too-late & \textvisiblespace too$\|$-l$\|$ittle$\|$-too$\|$-l$\|$ate & \textvisiblespace too$\|$-li$\|$ttl$\|$e$\|$-too$\|$-lat$\|$e & \textvisiblespace too$\|$-$\|$little$\|$-too$\|$-$\|$late \\
        even-valued & even$\|$-val$\|$ued & even$\|$-value$\|$d & even$\|$-value$\|$d \\
        \textvisiblespace eigenvectors & \textvisiblespace eig$\|$env$\|$ectors & \textvisiblespace eigen$\|$vector$\|$s & \textvisiblespace eigen$\|$vector$\|$s \\
        \textvisiblespace extrinsic & \textvisiblespace extr$\|$insic & \textvisiblespace extr$\|$insic & \textvisiblespace ex$\|$trinsic \\
        \textvisiblespace P(Queen & \textvisiblespace P$\|$(Q$\|$ue$\|$en & \textvisiblespace P$\|$(Q$\|$ue$\|$en & \textvisiblespace P$\|$($\|$Queen \\
        Combinatorics & Com$\|$bin$|$ator$\|$ics & Comb$\|$inator$\|$ics & Comb$\|$inator$\|$ics \\
        studentshare.org & students$\|$hare$\|$.org & students$\|$hare$|$.org & student$\|$share$\|$.org \\
        \}'.format & \}'.$\|$format & \}'.$\|$format & \}'$\|$.format \\
        (taxonomy & (t$\|$ax$\|$onomy & (t$\|$axon$\|$omy & ($\|$taxonomy \\
        AddIntegers & Add$\|$Int$\|$egers & Add$\|$Integer$\|$s & Add$\|$Integer$\|$s \\
        memory\_efficient & memory$\|$\_eff$\|$icient & memory$\|$\_eff$\|$icient & memory$\|$\_$\|$efficient \\
        \textvisiblespace pysimplesoap & \textvisiblespace pys$\|$imp$\|$les$\|$o$\|$ap & \textvisiblespace pys$\|$imple$\|$soap & \textvisiblespace py$\|$simple$\|$soap \\
        \bottomrule
    \end{tabular}
\end{table}

\begin{table}[t]
    \centering
    \caption{Example tokenization results of different encoding algorithms operating on the mix250k vocabulary}
    \label{tab:tokenization-examples-mix250k}
    \small
    \begin{tabular}{llll}
        \toprule
        Text & BPE & FMM & MCE \\
        \midrule
        \textvisiblespace were\textvisiblespace more\textvisiblespace likely\textvisiblespace to\textvisiblespace suffer & - & \textvisiblespace were\textvisiblespace more$\|$\textvisiblespace likely\textvisiblespace to$\|$\textvisiblespace suffer & \textvisiblespace were$\|$\textvisiblespace more\textvisiblespace likely\textvisiblespace to$\|$\textvisiblespace suffer \\
        \textvisiblespace that\textvisiblespace approximately & - & \textvisiblespace that\textvisiblespace a$\|$pp$\|$rox$\|$imate$\|$ly & \textvisiblespace that$\|$\textvisiblespace approximately \\
        \textvisiblespace move\textvisiblespace together & - & \textvisiblespace move\textvisiblespace to$\|$get$\|$her & \textvisiblespace move$\|$\textvisiblespace together\\
        Double\textvisiblespace Digit\textvisiblespace Addition & - & Double$\|$\textvisiblespace Dig$\|$it$\|$\textvisiblespace Addition & Double$\|$\textvisiblespace Digit$\|$\textvisiblespace Addition \\
        \textvisiblespace is\textvisiblespace to\textvisiblespace find\textvisiblespace the\textvisiblespace sum\textvisiblespace of\textvisiblespace the & - & is\textvisiblespace to$\|$\textvisiblespace find\textvisiblespace the$\|$\textvisiblespace sum\textvisiblespace of$\|$\textvisiblespace the & is$\|$\textvisiblespace to\textvisiblespace find$\|$\textvisiblespace the\textvisiblespace sum\textvisiblespace of\textvisiblespace the \\
        \bottomrule
    \end{tabular}
\end{table}

\clearpage
\section{Downstream Evaluation Details}
\label{app:downstream-evaluation}

\subsection{Model Configurations}
\label{app:model-configurations}

Table~\ref{tab:model-configurations} lists the configurations of our 1.8B- and 8B-scale CNF--MCE models used in Section~\ref{sec:exp-downstream}.

\begin{table}[h]
    \centering
    \caption{Configurations of the CNF--MCE models for downstream evaluation.}
    \label{tab:model-configurations}
    \small
    \begin{tabular}{lcc}
        \toprule
        Configuration & 1.8B & 8B \\
        \midrule
        Total parameters & 1.89B & 8.99B \\
        Non-embedding parameters & 1.61B & 6.95B \\
        Model dimension & 2048 & 4096 \\
        MLP hidden dimension & 6144 & 12288 \\
        Head dimension & 128 & 128 \\
        Number of query heads & 16 & 32 \\
        Number of key/value heads & 8 & 8 \\
        Number of layers & 32 & 36 \\
        Vocabulary size & 138112 & 249088 \\
        Tied embeddings & Yes & No \\
        \bottomrule
    \end{tabular}
\end{table}

\subsection{Benchmarks}
\label{app:downstream-benchmarks}

We summarize the 11 English benchmarks used in Section~\ref{sec:exp-downstream}. They cover scientific and commonsense reasoning, mathematics, code generation, contextual word prediction, and factual and subject knowledge. The shot settings are specified in the main text; evaluation methods and metrics are detailed in Appendix~\ref{app:downstream-metrics}.

\textbf{ARC}: A multiple-choice science QA dataset split into Challenge (hard) and Easy sets, testing grade-school level reasoning \citep{clark2018arc}.

\textbf{CommonsenseQA}: A multiple-choice dataset that requires commonsense knowledge, derived from ConceptNet, with questions that often require inferential reasoning beyond the surface text \citep{talmor2019commonsenseqa}.

\textbf{GSM8K}: A dataset of grade-school math word problems, testing multi-step arithmetic reasoning and language understanding \citep{cobbe2021gsm8k}.

\textbf{HellaSwag}: A commonsense NLI dataset where a model must choose the most plausible continuation for a given context, with adversarially filtered examples to avoid simple pattern matching \citep{zellers2019hellaswag}.

\textbf{HumanEval}: A code generation benchmark containing 164 handwritten programming problems with unit tests, used to evaluate functional correctness of generated code \citep{chen2021humaneval}.

\textbf{Lambada}: A cloze-style word prediction task in which models must predict the final word of a passage, requiring broader contextual understanding \citep{paperno2016lambada}.

\textbf{MMLU} (Massive Multitask Language Understanding): A broad multitask benchmark covering 57 subjects (e.g., math, history, law) in a multiple-choice format, designed for few-shot evaluation \citep{hendrycks2021mmlu}.

\textbf{OpenbookQA}: A multiple-choice QA dataset requiring reasoning over a small set of provided facts (an "open book"), testing the ability to combine knowledge with explanation \citep{mihaylov2018openbookqa}.

\textbf{PIQA} (Physical Interaction Question Answering): A multiple-choice QA dataset focused on physical commonsense reasoning, where models must choose the most sensible solution for everyday tool-use and object manipulation tasks, with questions designed to require real-world physical knowledge rather than language patterns \citep{bisk2020piqa}.

\textbf{TriviaQA}: A large-scale reading comprehension dataset with question-answer pairs from trivia websites, requiring fact retrieval and reasoning across documents \citep{joshi2017triviaqa}.

\textbf{Winogrande}: A scaled-up version of Winograd Schema challenge, a coreference resolution task that requires commonsense reasoning to resolve ambiguous pronouns \citep{sakaguchi2020winogrande}.

\subsection{Evaluation Methods and Metrics}
\label{app:downstream-metrics}

\paragraph{Accuracy under cloze prompting (CP).}
For ARC, CommonsenseQA, HellaSwag, MMLU, OpenBookQA, PIQA, and WinoGrande, we use cloze prompting. Each candidate answer is concatenated with the question or context to form a complete text sequence. We compute the model's perplexity for each sequence and select the candidate with the lowest perplexity (PPL). The reported score is accuracy, i.e., the percentage of examples for which the selected candidate matches the reference answer. Since base models have not been trained to follow instructions, multiple-choice prompting (MC) typically requires in-context examples to elicit option labels, whereas CP does not require such examples and thus supports zero-shot evaluation.

\paragraph{Exact match (EM).}
For GSM8K, LAMBADA, and TriviaQA, we report the percentage of predictions that exactly match an accepted reference answer after task-specific answer extraction and normalization.

\paragraph{Pass@1.}
For HumanEval, we report pass@1, which measures the probability that a single generated program passes all unit tests for a problem, averaged across problems.